\documentclass[runningheads]{llncs}
\usepackage[T1]{fontenc}
\usepackage{url}
\usepackage{flushend}
\usepackage{cite}
\usepackage{amsmath,amssymb,amsfonts}
\usepackage{algorithmic}
\usepackage{graphicx}
\usepackage{multirow}
\usepackage{textcomp}
\usepackage{float}
\usepackage{booktabs}
\usepackage{xcolor}
\usepackage[
  colorlinks=true,
  linkcolor=black,
  citecolor=black,
  urlcolor=black
]{hyperref}\usepackage{balance}
\begin{document}
\title{Exploring deep learning for Event-Based Saliency Prediction with a Transformer-based model}
\author{Romaric Mazna\inst{1}\orcidID{0009-0004-4658-0685} \and
Jean Martinet\inst{1}\orcidID{0000-0001-8821-5556} \and
Sai Deepesh Pokala\inst{1}\orcidID{0009-0008-2837-2988}}
\authorrunning{R. Mazna, J. Martinet, D. Pokala}
\titlerunning{Event-Based Saliency Prediction with a Transformer-based model}
\institute{\textsuperscript{1}i3S/CNRS,  Université Côte d'Azur\\
\email{\{firstname.surname\}@univ-cotedazur.fr}}
\maketitle              % typeset the header of the contribution

\begin{abstract}
Saliency prediction has been extensively studied in RGB images and videos as a computational model of human visual attention. In contrast, predicting saliency from event-based data remains largely unexplored, despite the biological inspiration and favorable sensing properties of event cameras. Two obstacles have held this direction back: the absence of large-scale event saliency datasets, and the lack of a strong baseline. In this paper, we introduce SEST (Swin Event-based Saliency Transformer), a transformer-based model for saliency prediction from event data, bridging the data scarcity barrier through event-native pretraining and synthetic supervision. SEST leverages a self-supervised pretrained event-based Swin Transformer backbone combined with a CNN decoder to produce dynamic saliency maps. To address the scarcity of annotated event-based saliency data, we introduce two new benchmark datasets, N-DHF1K and N-UCF Sports, generated from large-scale RGB saliency benchmarks. Experimental results show that SEST clearly outperforms existing event-based saliency methods and narrows the performance gap with state-of-the-art RGB models. Zero-shot evaluation on a real event camera dataset further demonstrates that our model trained on synthetic data remains transferable on real event streams. To the best of our knowledge, this work is the first to apply deep learning to event-based saliency prediction, opening a new research direction at the intersection of event-based vision and neuromorphic visual attention. \textit{Code and data are available at:  \href{https://github.com/Romageek/sest}{https://github.com/romageek/sest}}.

\keywords{Saliency \and Visual attention \and Event-based data \and Transformer}

%Saliency prediction is a long-studied domain. Several models have been designed to predict saliency from RGB images and videos, with the objective of mimicking human visual attention. Most applications of RGB saliency prediction target compression, summarization, and captioning. Yet, little attention has been given to predicting saliency from event-based data. Event-based data are a novel type of data generated by event cameras with many advantages such as high temporal resolution, low latency, and high dynamic range. 

%In this paper, we introduce SEST (Swin Event-based Saliency Transofmer), a novel model for predicting saliency from event-based data, based on a Swin Transformer as a backbone. We build a CNN decoder on top of this backbone to reconstruct feature maps into final saliency maps. The model has been trained on two new synthetic datasets harnessing human gaze data from RGB video. The results show that SEST can successfully predict saliency in event data even when trained with RGB-based groundtruth.
\end{abstract}    
\section{Introduction}
\label{sec:intro}

Visual attention refers to the ability of the human brain to selectively focus on the most relevant parts of a scene while ignoring less important details. Saliency prediction aims to model this phenomenon and has many applications in computer vision such as image captioning, image and video compression, and video summarization. Over the past years, several models have been developed to predict saliency in static and dynamic RGB data. Large-scale datasets with dense human fixation annotations have enabled the training of powerful convolutional, recurrent and transformer-based architectures capable of capturing complex spatial and temporal attention patterns. However, these advances rely only on visually dense input.

Recently, event cameras have seen a growing interest due to their advantages such as high dynamic range, low latency, energy efficiency and high temporal resolution. Contrary  to conventional RGB cameras, where images are recorded at a fixed rate and resolution, event cameras mimic the functioning of the human retina by detecting and recording brightness changes at pixel-level, thus removing redundancy in the generated data. While deep learning has been successfully applied to several event-based tasks such as object detection, optical flow and semantic segmentation, it has not yet addressed visual attention or saliency prediction.

Research on saliency prediction in event-based data is still at an early stage. In the context of RGB data, different architectures from low-level visual features  to deep learning methods have been developed to predict saliency. However, in the context of event-based data, the few models that have been developed are either hand-designed (proto-object models, per-pixel saliency pipelines) or use spiking architectures without gradient-based training. This can be explained by the lack of available human fixation datasets. Currently, only one event-based saliency dataset \cite{chane2024event} exists, comprising 598 clips of 8s each across a limited set of 6 imbalanced scene categories recorded with a stationary camera. Most of the few studies on saliency prediction in event-based data have relied on synthesizing a few samples of images/videos from RGB saliency datasets into events to evaluate their models. This lack of data has limited research in saliency prediction for event-based data.
We argue this barrier can be bridged by combining self-supervised event pretraining with synthetic supervision. We thus  leverage a Swin Transformer pretrained exclusively on event data \cite{yang2024event}, followed by a CNN decoder to produce saliency maps.

Furthermore, we introduce two new event-based saliency datasets, N-DHF1K and N-UCF Sports. We transform videos from video saliency datasets DHF1K \cite{wangRevisitingVideoSaliency2018} and UCF Sports \cite{mathe2014actions} into events using the ESIM library\cite{Gehrig_2020_CVPR}. The DHF1K dataset consists of 1K videos with more than 600K frames and per-frame fixation annotations from 17 observers across 150 scene categories covering diverse motions and activities. The UCF Sports dataset consists of 150 videos chosen from the UCF Sports Action dataset \cite{rodriguez2008action}. This scale and diversity enable far more robust saliency models than the only available real event-based dataset.

From a neuromorphic perspective, saliency prediction in event-based data aligns naturally with the principles of sparse, asynchronous, and temporally precise neural computation. Unlike frame-based saliency models that operate on dense visual representations, event-based saliency emerges from spatio-temporal contrast and motion cues, closely resembling bottom-up attentional mechanisms observed in biological vision. Although the proposed architecture is implemented using conventional deep learning components, it exploits event sparsity, temporal binning, and multi-scale temporal integration, making it compatible with future spiking adaptations for neuromorphic deployments.

The contributions of this work are the following:
\begin{enumerate}

%\item We introduce SEST, the first transformer-based deep learning architecture for saliency prediction in event-based data.
%\item We leverage self-supervised pretrained event transformers to model spatio-temporal saliency dynamics.
\item We present the first deep-learning approach to event-based saliency prediction, opening a research direction the field has not yet explored, and providing a baseline architecture (SEST) that the community can build on.
\item We introduce two event-based saliency datasets, N-DHF1K and N-UCF Sports, generated from large-scale RGB benchmarks to address the scarcity of annotated event saliency data.
\item We demonstrate that SEST significantly outperforms existing event-based saliency methods and transfers to real event streams in a zero-shot setting, validating the utility of our synthetic benchmarks.
\end{enumerate}
The paper is structured as follows.  Sec.~\ref{sec:rel_work} 
reviews related work, Sec.~\ref{sec:approach} details the proposed architecture, 
Sec.~\ref{sec:exps} describes the experimental setup, Sec.~\ref{sec:res_analys} presents and discusses the results, and Sec. ~\ref{sec:conclusion} concludes the paper.

\section{Related work}
\label{sec:rel_work}
%\subsection{Computational models for saliency prediction}
\subsection{Saliency prediction in RGB data} There have been various studies aimed at developing computational models for saliency prediction in RGB data. Early traditional approaches were based on the feature integration theory ~\cite{treisman1980feature}. The most commonly known work is the Itti and Koch model~\cite{itti2001computational}, which combines multi-scale image features into a single topographical saliency map. Le Meur et al. \cite{le2006coherent} refined this approach by incorporating additional human visual system (HVS) features, such as contrast sensitivity, perceptual decomposition, visual masking, and center-surround interactions. Other successful models include GBVS \cite{harel2006graph}, which constructs a fully connected graph over multiscale feature maps, and Hou et al. \cite{hou2008dynamic}, which introduces the Incremental Coding Length approach to maximize entropy gain in feature selection. Several static saliency models have also been proposed \cite{navalpakkam2006integrated, murray2011saliency, kootstra2008paying, gao2004discriminant, bruce2005saliency}, all relying on multi-scale feature computation to generate saliency maps. 
With advancements in deep learning, deep neural networks have been employed to predict saliency maps. Early CNN-based saliency models include EDN \cite{pan2016shallow} and DeepFix \cite{kruthiventi2017deepfix}, the latter incorporating VGG-16 features with location-based convolutional layers. Later approaches, such as SalGAN \cite{pan2017salgan}, leveraged GANs to enhance saliency predictions in static images through adversarial training, while EML-NET \cite{jia2020eml} introduced a disjoint encoder-decoder architecture. More recently, \cite{djilali2023vision} compared Vision Transformer (ViT) self-attention maps with saliency maps, demonstrating strong similarities, particularly under self-supervised training. With the availability of large video eye tracking datasets, several works focused on dynamic saliency modeling to leverage temporal context. Early deep video saliency models used convolutional architectures with recurrent units, including ConvLSTM-based architectures such as SalEMA \cite{linardos2019simple} and DeepVS\cite{Jiang_2018_ECCV}, and attentive ConvLSTM models like ACLNet \cite{wangRevisitingVideoSaliency2018} and SalSAC \cite{wu2020salsac}. Other works using convolutional architectures include TASED Net \cite{min2019tased}, HD2S\cite{bellitto2021hierarchical}, which use 3D CNNs to capture spatiotemporal features directly through volumetric convolutions, thereby avoiding recurrent units. 
Finally, more recent approaches increasingly rely on attention mechanisms and transformers to capture dependencies over a larger time-frame. Models such as STSANet \cite{Wang_2023} and STRA-Net\cite{lai2019video} introduce spatiotemporal self-attention and residual attentive modules to improve feature integration across space and time. Fully convolutional encoder-decoder architectures such as ViNet \cite{jain2021vinet}, TSFP\cite{chang2021temporalspatialfeaturepyramidvideo}, UNISAL \cite{Droste_2020} integrate multi-scale visual features for saliency prediction. Finally, transformer-based models including SalFoM \cite{moradi2024salfom}, TMFINet\cite{10130326zhou}, THTDNet\cite{ Moradi2024} further extend this approach by leveraging global context modeling, showing strong performance on complex dynamic scenes. \\
\subsection{Saliency prediction in event-based data} Few models have been developed in the context of event-based data. Iacono et al. (2019) \cite{iacono2019proto} adapted a proto-object attention model for event-based data to enhance iCub humanoid robot visual processing capabilities. Using a bottom-up attention mechanism, the model processes event-based data through three layers: center-surround filtering, border ownership cells, and grouping cells. These layers decompose the visual scene into proto-objects, with saliency determined by contrast and edges. D'Angelo et al. (2022) \cite{dangeloEventDrivenBioinspired2022} later implemented the proto-object attention model on SpiNNaker neuromorphic computing platform \cite{furber2020spinnaker}, effectively taking advantage of the asynchronous output of event cameras to reduce both latency and computational cost.
Studies by Gruel et al. in 2022 \cite{gruel2022neuromorphic} and
by Bulzomi et al. in 2023 \cite{bulzomi2023object} demonstrate the advantages of
neuromorphic attention for event-based data. Gruel's work focused on leveraging attention for gesture recognition, while Bulzomi applied attention to pedestrian detection using a static event camera. Both approaches employed a spiking neural network that dynamically adapts to incoming event-based data, focusing on regions of high activity as meaningful areas while filtering irrelevant background noise. More recently, Simon Chane et al. \cite{chane2024event} developed a pipeline used to compute a saliency score for each event as they occur. Their approach was tested on a dataset they captured from an event-based camera recording a dynamic visual scene. \\
%\subsection{Dataset limitations} While frame-based saliency benchmarks benefit from decades of eye-tracking data collection, resulting in large-scale datasets such as SALICON \cite{jiang2015salicon}, TORONTO \cite{bruce2007attention}, MIT300 \cite{judd2012benchmark} and CAT2000 \cite{borji2015cat2000} with dense spatial annotations at frame level, replicating the same process for event-based data is more challenging due to the asynchronous nature of event-based data. Collecting human annotations aligned with event-based data is a problem that has not been addressed as extensively as in the traditional frame-based context. To the best of our knowledge, there is only one event-based saliency prediction dataset that also has eye-tracking data from human observers\cite{chane2024event}. We thus address this by converting two RGB frame-based saliency datasets into the event-based domain. 

%%%%%%%%%%%%%%%%%%%%%%%%%%% MODELS DESCRIPTION %%%%%%%%%%%%%%%%%%%%%%%%%%%%
\section{Approach}
\label{sec:approach}

\begin{figure*}[ht]
    \centering
    \includegraphics[width=0.95\textwidth]{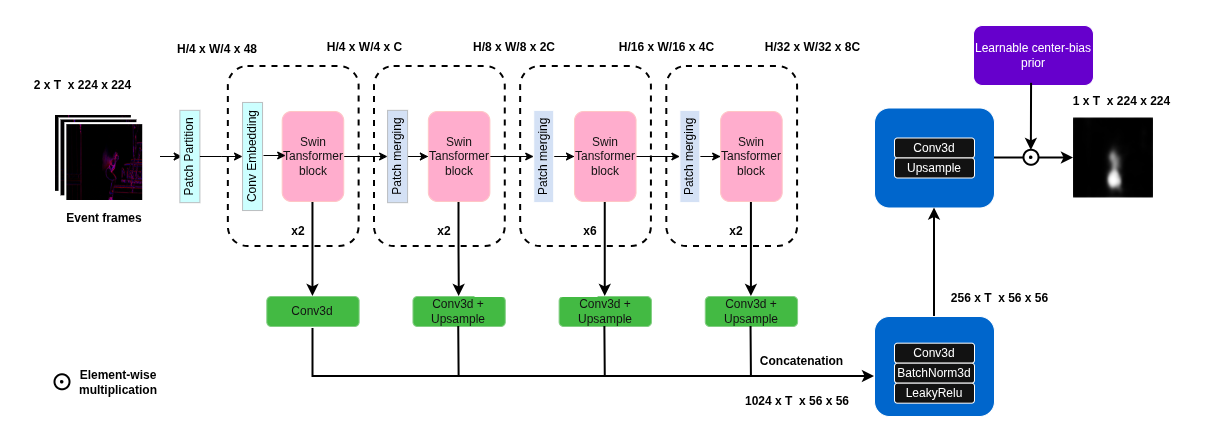}
    \caption{Overview of the proposed Swin Event-based Saliency Transformer (SEST) architecture.}
    \label{fig:architecture}
\end{figure*}
The proposed network (Figure \ref{fig:architecture}) is based on a pretrained Swin-T Transformer as encoder to extract multi-level spatiotemporal features from events input, followed by a CNN decoder to  decode features into saliency maps. We deliberately keep the architecture minimal to establish a clean baseline; methodological elaboration is left to future work.

\subsection{Event Representation}

We convert the raw event stream, represented as a set of asynchronous events 
$\{(x_k, y_k, t_k, p_k)\}$, into a voxel grid representation. Events within each temporal bin are accumulated per pixel and per polarity, producing a count-based voxel grid of shape [T, 2, H, W], H=W=224. During training, samples are processed in batches of size $B$, 
giving $\mathbf{X} \in \mathbb{R}^{B \times T \times 2 \times H \times W}$. 
This representation preserves both the temporal structure and the polarity 
asymmetry of the event stream.

\subsection{Pretrained Swin Transformer}

Yang et al.~\cite{yang2024event} introduced the first transformer solely trained 
with event-based data using a self-supervised scheme building on the DINOv2 framework. We leverage this pretrained backbone through end-to-end fine-tuning, inheriting rich event-specific representations learned from large-scale self-supervised training. Full pretraining details are described in \cite{yang2024event}.

%For each event input, a target view $\mathbf{x}^{+}$ and an augmented view $\mathbf{x}^{*}$ are generated and partitioned into $N$ patches. In the student network, a masking vector $\mathbf{m} \in \{0, 1\}^{N}$ is applied to $\mathbf{x}^{*}$, where $m_i = 1$ denotes a masked patch. The teacher network processes the unmasked view $\mathbf{x}^{+}$. To overcome the inherent sparsity of event data, the authors proposed the following objective:

%\begin{equation}
%    \mathcal{L}_{\text{total}} = \mathcal{L}_{\text{patch}} 
%    + \lambda_1 \mathcal{L}_{\text{context}} 
%    + \lambda_2 \mathcal{L}_{\text{image}}
%\end{equation}

%\noindent where $\mathcal{L}_{\text{patch}}$ aligns masked student patches with teacher outputs to learn local features, $\mathcal{L}_{\text{context}}$ aggregates features into $K$ context embeddings to capture contextual relationships between patches, and $\mathcal{L}_{\text{image}}$ ensures the alignment of global features between teacher and student networks. By leveraging this pretrained backbone, our model inherits event-specific representations learned from large-scale self-supervised training --- representations that neither a randomly initialised nor an RGB-pretrained backbone can provide.

\subsection{3D Feature Extraction and Fusion}

While the backbone is pretrained on 2D event images, dynamic saliency prediction 
requires reasoning over temporal sequences of event bins. To preserve the integrity of the 
pretrained 2D spatial representations, we treat each temporal bin as an 
independent spatial sample during encoding, while the temporal axis is explicitly 
recovered after feature extraction for subsequent 3D spatiotemporal reasoning.

Concretely, the batched input $\mathbf{X} \in \mathbb{R}^{B \times T \times 2 
\times H \times W}$ is reshaped into a flattened spatial batch $\mathbf{X}' \in 
\mathbb{R}^{(B \cdot T) \times 2 \times H \times W}$ before being fed into the 
encoder. This decoupling allows us to fully exploit the pretrained event 
representations without modification, while enabling the decoder to model 
cross-bin temporal dependencies that are essential for dynamic saliency 
prediction.

After processing, we extract hierarchical features from the four stages of the 
Swin-T Transformer. These features are reshaped back into 5D tensors 
$\{\mathbf{F}_i\}_{i=1}^{4}$, where $\mathbf{F}_i \in \mathbb{R}^{B \times T 
\times C_i \times H_i \times W_i}$ corresponds to the features from the 
$i^{\text{th}}$ Swin stage, restoring the temporal axis for subsequent 3D 
convolutional operations.

To align feature dimensions across scales, each stage output is processed by a 
$3 \times 3 \times 3$ Conv3D layer projecting channels to a fixed depth $d = 
256$. The 3×3×3 kernel captures local spatiotemporal context across adjacent temporal bins at each scale, compensating for the temporal independence assumed during the 2D encoding stage and introducing cross-bin awareness before fusion. Features from stages 2, 3, and 4 are spatially upsampled to $56 \times 
56$ via trilinear interpolation to match the resolution of stage 1:

\begin{equation}
    \mathbf{F}'_i = 
    \begin{cases} 
        \text{Conv3D}_i(\mathbf{F}_i), 
            & i = 1 \\[4pt] 
        \text{Upsample}\!\left(\text{Conv3D}_i(\mathbf{F}_i)\right), 
            & i \in \{2, 3, 4\} 
    \end{cases}
\end{equation}

\noindent The resulting features $\{\mathbf{F}'_i\}_{i=1}^{4}$ are concatenated 
along the channel dimension to form a unified spatiotemporal volume:

\begin{equation}
    \mathbf{U} = \text{Concat}\!\left(\{\mathbf{F}'_i\}_{i=1}^{4}\right) 
    \in \mathbb{R}^{B \times T \times 1024 \times 56 \times 56}
\end{equation}

\subsection{Feature Refinement and Saliency Map Reconstruction}

The concatenated volume $\mathbf{U}$ aggregates complementary information across 
four levels of abstraction: fine-grained boundary and texture information from 
stage 1 through to high-level semantic content at stage 4. A 3D convolutional 
refinement block integrates these representations jointly across channel and 
temporal dimensions:

\begin{equation}
    \mathbf{Z} = \text{LeakyReLU}\!\left(\text{BN}\!\left(
    \text{Conv3D}(\mathbf{U})\right)\right)
\end{equation}

\noindent Unlike a 2D convolution, the 3D kernel explicitly models how 
multi-scale features co-evolve across time, capturing for instance that a 
salient object boundary at temporal bin $t$ should remain spatially consistent 
with the semantic saliency response at the same location in adjacent bins. The 
channel dimension is simultaneously reduced from 1024 to $d = 256$, into a compact representation $\mathbf{Z} \in \mathbb{R}^{B \times T 
\times 256 \times 56 \times 56}$.

The refined features are projected to the target channel depth and upsampled to 
the original input resolution to produce a dense prediction:

\begin{equation}
    \hat{\mathbf{Y}} = \text{Upsample}\!\left(\text{Conv3D}(\mathbf{Z})\right),
    \qquad \hat{\mathbf{Y}} \in \mathbb{R}^{B \times T \times 1 \times H \times W}
\end{equation}

To model spatial attention bias, we follow~\cite{cornia2016deep} and allow the 
network to learn a center-bias prior directly from training data rather than 
imposing a pre-defined prior. A learnable weight matrix $\mathbf{M}_b \in 
\mathbb{R}^{H \times W}$, initialized from a uniform distribution $\mathbf{M}_b 
\sim \mathcal{U}(0, 1)$, modulates the dense prediction via element-wise 
multiplication.

\begin{equation}
    \hat{\mathbf{Y}}_{\text{biased}} = \hat{\mathbf{Y}} \odot 
    \left(1 + \mathbf{M}_b\right)
\end{equation}

%\noindent where $\odot$ denotes element-wise multiplication. 
By learning this bias from data rather than assuming a 
fixed Gaussian prior, the model adapts to the statistical regularities of event-based data specifically.

The final saliency map is obtained by applying Gaussian blur smoothing followed 
by a Sigmoid activation.

%\begin{equation}
%    \hat{\mathbf{Y}}_{\text{final}} = 
%    \text{Sigmoid}\!\left(\text{GaussianBlur}\!\left(
%    \hat{\mathbf{Y}}_{\text{biased}}\right)\right)
%\end{equation}

%%%%%%%%%%%%%%%%%%%%%%%%%%% EXPERIMENTS %%%%%%%%%%%%%%%%%%%%%%%%%%%%%%%%%%%
\section{Experiments}
\label{sec:exps}
%%%%%%%%%%%%%%%%%%%%%%%%%%% DATASETS %%%%%%%%%%%%%%%%%%%%%%%%%%%%%%%%%%%
\subsection{Saliency datasets}

 We converted two existing video saliency datasets (DHF1K \cite{wangRevisitingVideoSaliency2018}, and UCF Sports \cite{mathe2014actions}) into event datasets (N-DHF1K and N-UCF Sports) using ESIM library\cite{Gehrig_2020_CVPR} with positive and negative contrast thresholds of 0.09 and a 3 ms refractory period. We used their saliency and fixation maps as ground truths to train our model on the synthetic events.

\subsubsection{DHF1K}
Dynamic Human Fixations 1K (DHF1K) is a large-scale dataset of gaze recordings from 17 observers who were asked to freely view the videos. It consists of 1K videos with diverse content, varied motion and various objects. The original dataset was partitioned into 600 training, 100 validation, and 300 test videos. Since the ground truth labels for the official test are hidden for benchmarking, we reserved 100 videos from the training set to serve our test set. This resulted in a final split of 500 training videos, 100 validation videos, and 100 test videos.
\subsubsection{UCF Sports}
UCF Sports is a human fixations dataset on 150 videos from the UCF Sports Action dataset. The videos covered 9 action sports such as swinging, diving, and lifting. Contrary to DHF1K, where observers freely observed the videos, observers were asked to identify occurring actions within the UCF videos. Our final data split consists of 103 training videos, 15 validation videos, and a test set of 32 videos.

Although the proposed datasets are synthetically generated from RGB videos, this approach enables scalable supervision using high-quality human fixation data that is currently scarce in native event-based form. More importantly, prior work has shown that event representations generated from photorealistic videos preserve motion, contrast, and temporal saliency cues relevant to downstream tasks \cite{Gehrig_2020_CVPR, yang2024event}. 

\subsection{Protocol}
\subsubsection{Implementation details}
SEST is implemented using the PyTorch framework. We use a batch size of 20 and a learning rate of 0.006. We train our model on a single NVIDIA H100 GPU for 30 epochs with an AdamW optimizer \cite{loshchilovdecoupled}. We implement early stopping based on the validation loss (patience = 3) and reduce the learning rate on plateau by a factor of 0.1 (patience = 1). We trained and evaluated our model using various temporal window sizes by varying the number of bins ($T \in \{7, 10, 14, 21\}$) across the two datasets. The duration of each bin was fixed to match the sampling period of the original source videos. For N-UCF Sports (10Hz), the bin duration was set to 100ms, while for N-DHF1K (30Hz), it was set to 33.33ms. Consequently, the total observation window for each experiment scales linearly with the number of bins.

\subsubsection{Loss function}
We considered different metrics to design our loss function, following common practice in the literature \cite{bylinskiiWhatDifferentEvaluation2019}. Let $Y$ denote the ground truth saliency map and $\hat{Y}$ the predicted saliency map. The loss function is defined as: 
\begin{equation}
\mathcal{L}(\hat{Y}, Y)= \mathcal{L}_{\mathrm{KL}}(\hat{Y}, Y) + \alpha_1\mathcal{L}_{\mathrm{CC}}(\hat{Y}, Y) + \alpha_2\mathcal{L}_{\mathrm{BCE}}(\hat{Y}, Y)
\end{equation}
 with $\alpha_1 = 0.5$ and $\alpha_2 = 0.7$ set empirically. $ \mathcal{L}_{\mathrm{KL}}$, $\mathcal{L}_{\mathrm{CC}}$ and $\mathcal{L}_{\mathrm{BCE}}$ are \textit{Kullback-Leibler divergence}, \textit{Pearson Correlation Coefficient}, and \textit{Binary Cross Entropy} respectively. $ \mathcal{L}_{\mathrm{KL}} $ and $\mathcal{L}_{\mathrm{CC}}$ supervise the predicted saliency at the distribution level, while $\mathcal{L}_{\mathrm{BCE}}$ enforces pixel-wise localization, balancing distribution-matching with spatial accuracy. We negate the Pearson correlation so that minimizing $\mathcal{L}_{\mathrm{CC}}$ maximizes the correlation between $Y$ and $\hat{Y}$.

$ \mathcal{L}_{\mathrm{KL}} $ implements the \textit{Kullback-Leibler Divergence}, which quantifies the dissimilarity between two probability distributions:\\
\begin{equation}
\mathcal{L}_{\mathrm{KL}}(\hat{Y}, Y) = \sum_{x} {Y(x)} \log( \frac{Y(x)}{\hat{Y}(x)})
\end{equation}

$\mathcal{L}_{\mathrm{CC}}$ implements the \textit{Pearson Correlation Coefficient} that measures the linear relationship between two images:

\begin{equation}
\mathcal{L}_{\mathrm{CC}}(\hat{Y},Y) = - \frac{\sum_{i} (\hat{Y}_i - \bar{\hat{Y}})(Y_i - \bar{Y})}{\sqrt{\sum_{i} (\hat{Y}_i - \bar{\hat{Y}})^2} \sqrt{\sum_{i} (Y_i - \bar{Y})^2}}
\end{equation} 

$\mathcal{L}_{\mathrm{BCE}}$ implements the \textit{Binary Cross Entropy } that measures the pixel-wise dissimilarity between two images:
\begin{equation}
\mathcal{L}_{\mathrm{BCE}}(\hat{Y},Y) = - \sum_{i} \left[ Y_i \log(\hat{Y}_i) + (1 - Y_i) \log(1 - \hat{Y}_i) \right]
\end{equation}

\subsubsection{Evaluation metrics}
We evaluate our model performance by employing the following metrics: AUC-Judd (AUC-J), Pearson Correlation Coefficient (CC), Similarity (SIM), and Normalized Scanpath Saliency (NSS). These metrics have been discussed in detail in \cite{bylinskiiWhatDifferentEvaluation2019}. 

\section{Results and Analysis}
\label{sec:res_analys}
\subsection{Performance results}
We compared the saliency prediction performance of our event-based model, SEST, on the N-DHF1K and N-UCF Sports datasets to the performance of two state-of-the-art event-based models (SNNevProto \cite{dangeloEventDrivenBioinspired2022} and evST \cite{chane2024event}) and several state-of-art RGB-based models from DHF1K and UCF Sports benchmarks\footnote{\href{https://mmcheng.net/videosal}{https://mmcheng.net/videosal}} (see Table \ref{tab:comparison}). 

\begin{figure*}[ht]
    \centering
    \includegraphics[width=0.82\textwidth]{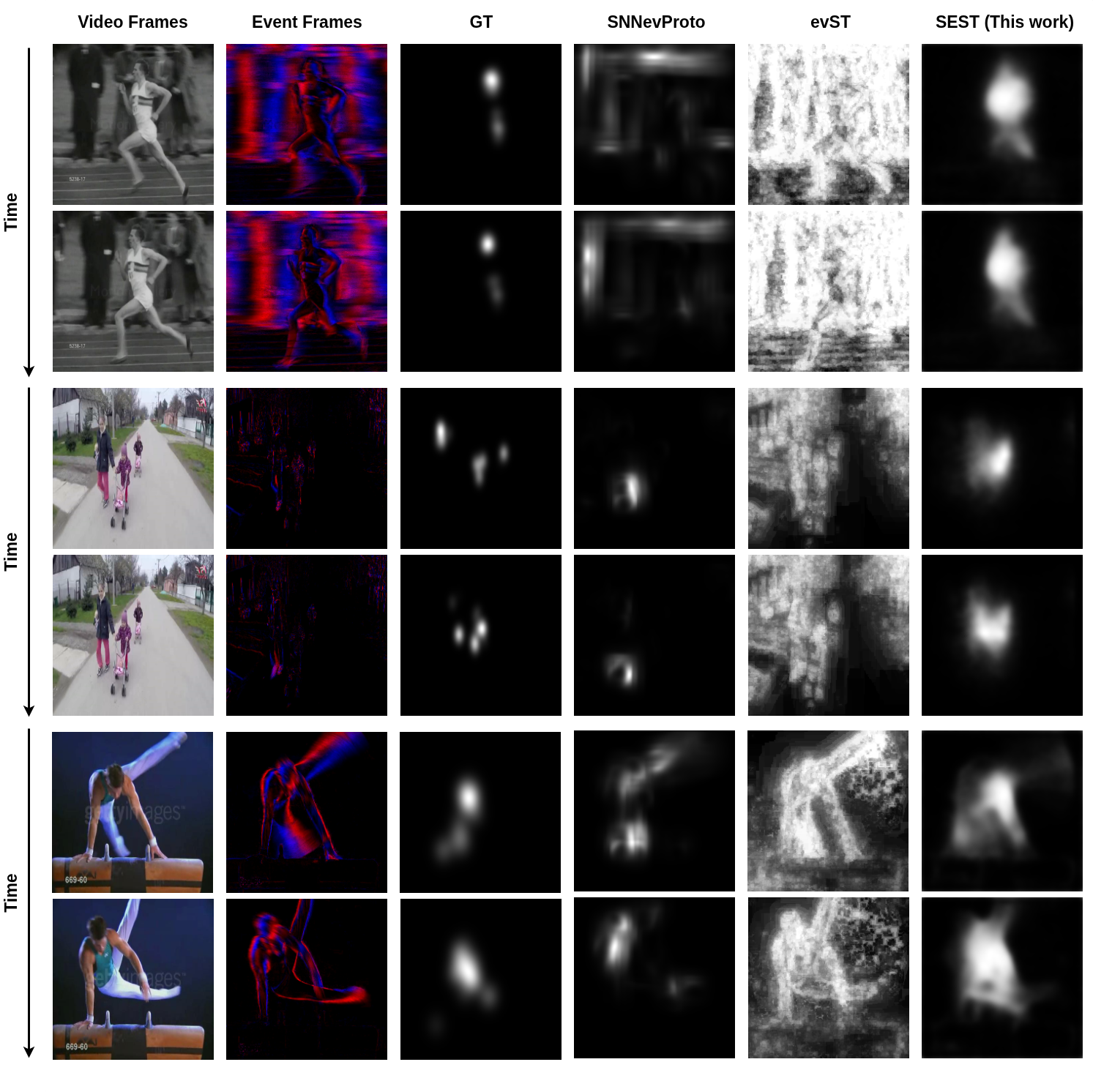}
    \caption{Illustration of qualitative results for three samples: UCF Sports Run-side-004 (rows 1-2), DHF1K  548 (rows 3-4), UCF Sports Swing-Bench-003 (rows 5-6).}
    \label{fig:viz}
\end{figure*}

\subsubsection{N-DHF1K} The best overall performance on this dataset is achieved by RGB-based models, with SalFoM leading across all four metrics. Other RGB-based models such as TMFI, THTD-Net, and STSANet follow closely. In contrast, existing event-based models show significantly inferior performance, with very low CC and NSS values. Our model, SEST, substantially improves performance within the event domain. Across different time bin configurations, SEST achieves AUC-J values up to 0.9197, CC up to 0.4907, SIM up to 0.3969, and NSS up to 2.3956, clearly outperforming all other event-based approaches by a large margin. 

\subsubsection{N-UCF Sports} Similar to the case of DHF1K, RGB models achieve the highest overall performance on this dataset too. Among them, the best performance is generally achieved by STSANet and TMFI, with AUC-J values of 0.936 and strong NSS scores. In comparison, existing event-based models exhibit noticeably lower performance, with AUC-J values typically in the range of 0.77 to 0.81 and NSS scores mostly below 1.5, highlighting the gap that remains between RGB-based methods and current event-based methods. However, SEST consistently outperforms all other event-based models across all four metrics. In particular, our model attains the highest AUC-J value among the event-based approaches, along with improved CC, SIM and NSS scores. Notably, the 14-bin configuration of our model provides the best overall performance. 

These results show that although a performance gap still exists between RGB and event-based methods, SEST significantly narrows this gap and sets a new state of the art for event-based saliency prediction on the N-DHF1K and N-UCF Sports datasets.

\begin{table*}[!hbt]
    \centering
    \caption{Quantitative comparison on N-DHF1K and N-UCF Sports datasets.}
    \label{tab:comparison}
    \renewcommand{\arraystretch}{1.2}
    \setlength{\tabcolsep}{5pt}
    \resizebox{\textwidth}{!}{
        \begin{tabular}{llccccc|llccccc}
            \toprule
            \multicolumn{7}{c|}{\textbf{N-DHF1K}} & \multicolumn{7}{c}{\textbf{N-UCF Sports}} \\
            \midrule
            \textbf{Domain} & \textbf{Model} & \textbf{\#bins} & \textbf{AUC-J}$\uparrow$ & \textbf{CC}$\uparrow$ & \textbf{SIM}$\uparrow$ & \textbf{NSS}$\uparrow$ &
            \textbf{Domain} & \textbf{Model} & \textbf{\#bins} & \textbf{AUC-J}$\uparrow$ & \textbf{CC}$\uparrow$ & \textbf{SIM}$\uparrow$ & \textbf{NSS}$\uparrow$ \\
            \midrule
            \multirow{5}{*}{RGB}
            & \textbf{SalFoM}~\cite{moradi2024salfom}        & - & \textbf{0.9222} & \textbf{0.5692} & \textbf{0.4208} & \textbf{3.3536} &
            \multirow{5}{*}{RGB}
            & \textbf{TMFI}~\cite{10130326zhou}              & - & \textbf{0.936} & \textbf{0.707} & \textbf{0.565} & 3.863 \\
            & \textbf{TMFI}~\cite{10130326zhou}              & - & 0.9153 & 0.5461 & 0.4068 & 3.1463 &
            & \textbf{STSANet}~\cite{Wang_2023}              & - & \textbf{0.936} & 0.705 & 0.560 & \textbf{3.908} \\
            & \textbf{THTD-Net}~\cite{Moradi2024}            & - & 0.9152 & 0.5479 & 0.4062 & 3.1385 &
            & \textbf{SalSAC}~\cite{wu2020salsac}            & - & 0.926 & 0.671 & 0.534 & 3.523 \\
            & \textbf{STSANet}~\cite{Wang_2023}              & - & 0.9125 & 0.5288 & 0.3829 & 3.0103 &
            & \textbf{ViNet}~\cite{jain2021vinet}            & - & 0.924 & 0.673 & 0.522 & 3.620 \\
            & \textbf{TSFP-Net}~\cite{chang2021temporalspatialfeaturepyramidvideo} & - & 0.9116 & 0.5168 & 0.3921 & 2.9665 &
            & \textbf{TSFP-Net}~\cite{chang2021temporalspatialfeaturepyramidvideo} & - & 0.923 & 0.685 & 0.561 & 3.698 \\
            \midrule
            \multirow{9}{*}{Event}
            & \textbf{evST}~\cite{chane2024event}            & 21 & 0.5464 & 0.0274 & 0.1106 & 0.1420 &
            \multirow{9}{*}{Event}
            & \textbf{evST}~\cite{chane2024event}            & 21 & 0.7756 & 0.2681 & 0.1464 & 1.3312 \\
            & \textbf{evST}~\cite{chane2024event}            & 14 & 0.5408 & 0.0232 & 0.1208 & 0.1091 &
            & \textbf{evST}~\cite{chane2024event}            & 14 & 0.7779 & 0.2658 & 0.1382 & 1.3692 \\
            & \textbf{evST}~\cite{chane2024event}            & 10 & 0.5413 & 0.0246 & 0.1290 & 0.1120 &
            & \textbf{evST}~\cite{chane2024event}            & 10 & 0.7793 & 0.2594 & 0.1308 & 1.3798 \\
            & \textbf{evST}~\cite{chane2024event}            &  7 & 0.5380 & 0.0217 & 0.1317 & 0.0898 &
            & \textbf{evST}~\cite{chane2024event}            &  7 & 0.7787 & 0.2559 & 0.1263 & 1.3798 \\
            & \textbf{SNNevProto}~\cite{dangeloEventDrivenBioinspired2022} & 21 & 0.6427 & 0.0753 & 0.1092 & 0.3999 &
            & \textbf{SNNevProto}~\cite{dangeloEventDrivenBioinspired2022} & 21 & 0.7890 & 0.2325 & 0.1336 & 1.0205 \\
            & \textbf{SNNevProto}~\cite{dangeloEventDrivenBioinspired2022} & 14 & 0.6201 & 0.0579 & 0.1185 & 0.2886 &
            & \textbf{SNNevProto}~\cite{dangeloEventDrivenBioinspired2022} & 14 & 0.8195 & 0.2741 & 0.1369 & 1.2204 \\
            & \textbf{SNNevProto}~\cite{dangeloEventDrivenBioinspired2022} & 10 & 0.6036 & 0.0443 & 0.1266 & 0.2054 &
            & \textbf{SNNevProto}~\cite{dangeloEventDrivenBioinspired2022} & 10 & 0.7941 & 0.2237 & 0.1144 & 1.0474 \\
            & \textbf{SNNevProto}~\cite{dangeloEventDrivenBioinspired2022} &  7 & 0.5884 & 0.0375 & 0.1342 & 0.1725 &
            & \textbf{SNNevProto}~\cite{dangeloEventDrivenBioinspired2022} &  7 & 0.8316 & 0.2834 & 0.1262 & 1.3286 \\
            & \textbf{SEST (this work)}                                  & 21 & 0.8956 & 0.4284 & 0.3014 & 2.1209 &
            & \textbf{SEST (this work)}                                  & 21 & 0.8605 & 0.4381 & 0.3090 & 2.1259 \\
            & \textbf{SEST (this work)}                                  & 14 & 0.9009 & 0.4814 & 0.3641 & 2.2769 &
            & \textbf{SEST (this work)}                                  & 14 & \textbf{0.8943} & \textbf{0.5237} & \textbf{0.3800} & \textbf{2.6590} \\
            & \textbf{SEST (this work)}                                  & 10 & 0.8947 & \textbf{0.4907} & \textbf{0.3969} & 2.2328 &
            & \textbf{SEST (this work)}                                  & 10 & 0.8942 & 0.5073 & 0.3462 & 2.6022 \\
            & \textbf{SEST (this work)}                      &  7 & \textbf{0.9197} & 0.4661 & 0.3634 & \textbf{2.3956} &
            & \textbf{SEST (this work)}                      &  7 & 0.8841 & 0.4555 & 0.3214 & 2.4537 \\
            \bottomrule
        \end{tabular}
    }
\end{table*}

\subsubsection{Qualitative analysis}
Figure~\ref{fig:viz} presents qualitative comparisons across three representative sequences: a high-motion athletics scene, a low-motion pedestrian scene, and a gymnastics scene. In all three cases, SEST produces spatially coherent, temporally consistent saliency maps that closely match the ground truth. SNNevProto generates diffuse activations dominated by background structure, failing to isolate salient subjects, while evST produces noisy, unstructured responses that deteriorate further in high-activity scenes. These observations highlight two consistent advantages of SEST over prior methods: robustness to varying event density, and temporal consistency across bins, both essential properties for reliable event-based saliency prediction in practice.

\subsubsection{Generalizability and Transferability on real event dataset}
We evaluated the cross-dataset generalizability of SEST in two complementary settings (Table~\ref{tab:generalizability_tab}). First, we assess synthetic-to-synthetic transfer by training on one synthetic dataset and testing on the other. The results show that training on N-DHF1K generalizes better to N-UCF Sports than the other way around. This suggests that N-DHF1K, which contains more diverse scenes and motion patterns, provides a richer training distribution that transfers more effectively to N-UCF Sports. Conversely, the more domain-specific sports content in N-UCF Sports leads to reduced generalization when evaluated on the broader N-DHF1K dataset.

Second, to assess generalization beyond synthetic data, we conduct a zero-shot evaluation on 100 sequences (sampled at 30Hz, equally distributed across the 6 scene categories) from the real event camera dataset of~\cite{chane2024event}, using 14 temporal bins and no fine-tuning. Despite the domain shift, SEST remains competitive: while evST obtains a higher NSS (1.4141), SEST achieves substantially higher CC and SIM (0.4725 and 0.4708), confirming generalizable saliency distributions. SNNevProto, in contrast, fails to generalise on AUC-J, CC, and NSS. SEST's lower NSS likely reflects its sensitivity to fixation-level spatial displacement under domain shift, and the low-motion nature of the real dataset may favor evST's asynchronous per-pixel design. This highlights a limitation of the current real dataset and motivates N-DHF1K and N-UCF Sports as essential resources, given their broader motion dynamics, for training robust event-based saliency models.

\begin{table}[H]
    \centering
    \caption{Cross-dataset generalization. The top rows report synthetic-to-synthetic transfer between N-UCF Sports and N-DHF1K. The bottom rows report zero-shot evaluation on the real event camera dataset ~\cite{chane2024event} (no fine-tuning).}
    \label{tab:generalizability_tab}
    \renewcommand{\arraystretch}{1.2}
    \setlength{\tabcolsep}{6pt}
    \resizebox{0.6\textwidth}{!}{
        \begin{tabular}{llcccc}
            \toprule
            \textbf{Model (Train / Test)} & \textbf{\#bins} & \textbf{AUC-J} $\uparrow$ & \textbf{CC} $\uparrow$ & \textbf{SIM} $\uparrow$ & \textbf{NSS} $\uparrow$ \\
            \midrule
            \multicolumn{6}{l}{\textit{Synthetic-to-synthetic transfer}} \\
            \textbf{SEST (N-UCF Sports / N-DHF1K)} & 14 & 0.7831 & 0.2598 & 0.2246 & 1.2331 \\
            \textbf{SEST (N-DHF1K / N-UCF Sports)} & 14 & 0.9015 & 0.5286 & 0.4256 & 2.6873 \\
            \midrule
            \multicolumn{6}{l}{\textit{Zero-shot on real event data~\cite{chane2024event}}} \\
            \textbf{SNNevProto ~\cite{dangeloEventDrivenBioinspired2022} (- / real event data)} & 14 & 0.5763 & 0.0480 & 0.3059 & 0.1126 \\
            \textbf{evST ~\cite{chane2024event} (-/ real event data)} & 14 & \textbf{0.7848} & 0.2750 & 0.1396 & \textbf{1.4141} \\
            \textbf{SEST (N-DHF1K/ real event data)} & 14 & 0.7725 & \textbf{0.4725} & \textbf{0.4708} & 1.1434 \\
            \bottomrule
        \end{tabular}
    }
\end{table}

\subsection{Computational Load}
We evaluated the inference time for a single saliency map prediction across all three models on both GPU and CPU (Table~\ref{tab:computational_tab}), using an NVIDIA RTX A5500 (16GB) and an Intel Core i9-12950HX. Despite its larger memory footprint, SEST achieves the fastest inference on both hardware configurations (6.62ms GPU, 121.68ms CPU). The high inference times of competing methods reflect their differing design goals: SNNevProto was optimized for the SpiNNaker neuromorphic platform~\cite{furber2020spinnaker}, where it achieves 16ms ~\cite{dangeloEventDrivenBioinspired2022}, while evST was designed for per-pixel saliency updates at microsecond precision rather than dense map reconstruction~\cite{chane2024event}, making direct latency comparison with frame-based approaches inherently limited. Table~\ref{tab:computational_tab} should therefore be read as a characterization of SEST's computational profile on general-purpose hardware, not as a ranking of overall efficiency.

%\begin{table}[hbt!]
%    \centering
%    \renewcommand{\arraystretch}{1.2}
%    \setlength{\tabcolsep}{6pt}
%    \resizebox{0.48\textwidth}{!}{
%        \begin{tabular}{llcccc}
%            \toprule
%            \textbf{Model} & \textbf{# bins} & \textbf{ GPU Inference time (ms)}  & \textbf{ CPU Inference time (ms)}  & \textbf{Model Size (MB)}\\
%            \midrule
%            \textbf{D'Angelo } & 14 & 877.7 & \textbf{19546.5} & - \\ 
%            \textbf{Simon Chane} & 14 & - & \textbf{2109.68} & - \\
%            \textbf{Ours} & 14 & 92.7 & 1703.6s & 180.247 \\
%
%            \bottomrule     
%        \end{tabular}
% }
%    \caption{Computational load(Results yet to be updated)}
%    \label{tab:computational_tab}
%\end{table}

\begin{table}[H]
    \centering
    \caption{Comparative inference times.}
    \label{tab:computational_tab}
    \renewcommand{\arraystretch}{1.2}
    \setlength{\tabcolsep}{6pt}
    \resizebox{0.48\textwidth}{!}{
        \begin{tabular}{llcccc}
            \toprule
            \textbf{Model}  & \textbf{ GPU Inference time (ms)}  & \textbf{ CPU Inference time (ms)}  & \textbf{Parameters (M)} & \textbf{Model size (MB)}\\
            \midrule
            \textbf{SNNevProto \cite{dangeloEventDrivenBioinspired2022}} &  62.69 & 1396.17 & -  & - \\ 
            \textbf{evST \cite{chane2024event}} & - & 2109.68 & -  & - \\
            \textbf{SEST (this work)} &  \textbf{6.62} & \textbf{121.68} & 45.1 & 180.247 \\
            \bottomrule    
        \end{tabular}
 }
\end{table}

\subsection{Ablation study}
In this section, we assess the impact of two design choices on the N-UCF Sports test set: the learnable center bias component and the use of 2D convolutions in place of 3D convolutions within the decoder.

\subsubsection{Training without center bias learning} To evaluate the contribution of the spatial prior, we conducted an ablation study by training the architecture without the learnable center-bias component across all temporal configurations. The comparative results are detailed in Table \ref{tab:no_center_bias}. Our analysis indicates that the effectiveness of the center-bias is highly dependent on the temporal window length. At moderate temporal windows (10-14 bins), the center-bias significantly improves the CC and NSS metrics. We attribute this improvement to the module’s implementation as a multiplicative spatial gain, $\hat{Y} \odot (1 + \mathbf{M}_b)$, which allows the network to adaptively amplify salient features within statistically probable regions of interest.

However, at 21 bins, the effectiveness  diminishes.  We interpret this as a conflict between the learned global spatial prior and the increased local variance of longer sequences; as the temporal window expands, the salient target is more likely to deviate from the statistically learned center-bias regions, leading to a suppression of valid features. These results suggest the center-bias is most effective for short-to-medium temporal integration.

\subsubsection{Replacing Conv3d with Conv2d}: We also trained a variant of the model in which all 3D convolutions in the decoder are replaced by 2D convolutions, deferring the recovery of the temporal axis to the very last layer. When trained on 14 bins, this variant yields degraded performance (AUC-J: 0.8483, CC: 0.4072, SIM: 0.2415, NSS: 1.9544), as expected, since Conv2d treats each time bin independently and cannot exploit temporal correlations until the output. Reducing the input to a single bin provides only a marginal improvement (AUC-J: 0.8410, CC: 0.4179, SIM: 0.2946, NSS: 2.0522), still falling short of the Conv3d baseline. These results indicate that 2D convolutions are not well suited to the task given our pretrained backbone and underline the importance of explicitly modeling the temporal information of event-based data, a property that future work should further exploit to improve performance.

\begin{table}[H]
    \centering
    \caption{SEST performance on N-UCF Sports without center bias learning.}
    \label{tab:no_center_bias}
    \renewcommand{\arraystretch}{1.2}
    \setlength{\tabcolsep}{6pt}
    \resizebox{0.48\textwidth}{!}{
        \begin{tabular}{llcccc}
            \toprule
            \textbf{Model} & \textbf{\#bins} & \textbf{AUC-J} $\uparrow$ & \textbf{CC} $\uparrow$ & \textbf{SIM} $\uparrow$ & \textbf{NSS} $\uparrow$ \\
            \midrule
            SEST (No Center bias) & 21 & 0.8954 & 0.5190 & 0.3498 &  2.6149\\
            SEST (No Center bias) & 14 & 0.8847 & 0.4978 & 0.3795 &  2.5049\\
            SEST (No Center bias) & 10 & 0.8861 & 0.4799 & 0.3516 &  2.4906\\
            SEST (No Center bias) & 7 & 0.8999 & 0.4908 & 0.3595 &  2.4978 \\
            \bottomrule     
        \end{tabular}
 }
\end{table}

\section{Conclusion} 
\label{sec:conclusion}
In this work, we addressed the largely unexplored problem of saliency prediction from event-based data. We proposed SEST, a transformer-based approach that leverages self-supervised pretrained event representations to model spatiotemporal saliency dynamics directly from event streams. To support supervised training and evaluation, we introduced two new event-based saliency datasets, N-DHF1K and N-UCF Sports, derived from large-scale RGB saliency benchmarks. While synthetically generated, these datasets enable scalable experimentation across diverse motion patterns and scene dynamics. Experimental results demonstrate that SEST significantly outperforms existing event-based saliency methods and narrows the performance gap with state-of-the-art RGB models. Zero-shot evaluation on a real event camera dataset further confirms that our synthetic benchmarks provide a transferable supervisory signal for real-world generalization.

Despite these encouraging results, the current architecture leverages temporal information only late in the decoder and remains too heavy for edge applications where event cameras typically excel. Future work will focus on explicit temporal attention over event bins and on lighter models through architectural redesign and compression techniques. By showing deep learning is viable for this task, we hope to open a new research direction at the intersection of event-based vision, saliency, and neuromorphic computing.

\begin{credits}
\subsubsection{\ackname} This work was supported by the project NAMED (ANR-23-CE45-0025-01) of the French National Research Agency (ANR). Training experiments presented in this paper were carried out using the Grid’5000 testbed, supported by a scientific interest group hosted by Inria and including CNRS, RENATER and several Universities as well as other organizations (see \href{https://www.grid5000.fr}{https://www.grid5000.fr}).
%\subsubsection{\discintname}
%It is now necessary to declare any competing interests or to specifically state that the authors have no competing interests. Please place the statement with a bold run-in heading in small font size beneath the (optional) acknowledgments\footnote{If EquinOCS, our proceedings submission system, is used, then the disclaimer can be provided directly in the system.}, for example: The authors have no competing interests to declare that are relevant to the content of this article. Or: Author A has received research grants from Company W. Author B has received a speaker honorarium fromCompany X and owns stock in Company Y. Author C is a member of committee Z.
\end{credits}
 \bibliographystyle{splncs04}
 \bibliography{main}

\end{document}